\documentclass[letterpaper, 10 pt, conference]{ieeeconf}  

\IEEEoverridecommandlockouts                              

\usepackage[backend=bibtex,style=ieee,natbib=true,mincitenames=1,maxcitenames=2]{biblatex}
\usepackage{algorithm2e}
\usepackage{amsmath} 
\usepackage{amssymb}  
\usepackage{array}
\usepackage{comment}
\usepackage{epsfig} 
\usepackage{graphicx} 
\usepackage{hyperref}
\usepackage{stmaryrd}

\usepackage[dvipsnames]{xcolor}

\newcommand{\VA}[1]{\textcolor{green}{Varun: #1}}

\title{\LARGE \bf
Continuous-time State \& Dynamics Estimation \\
using a Pseudo-Spectral Parameterization
}

\author{Varun Agrawal$^{1}$ and Frank Dellaert$^{1}$
\thanks{$^{1}$The authors are affiliated with the Institute for Robotics and Intelligent Machines,
            Georgia Institute of Technology, 
            Atlanta GA 30332, USA.
        {\tt\small email: \{varunagrawal,frank.dellaert\}@gatech.edu}}%
}

\begin{document} 

\noindent

\maketitle
\thispagestyle{empty}
\pagestyle{empty}

\global\long\def\Vector#1{{\bf #1}}
 \global\long\def\Matrix#1{{\bf #1}}

\global\long\def\eq#1{equation (\ref{eq:=0000231})}

\global\long\def\eye#1{\Vector{I_{#1}}}

\global\long\def\leftsparrow#1{\stackrel{#1}{\leftarrow}}
 \global\long\def\rightsparrow#1{\stackrel{#1}{\rightarrow}}
 \global\long\def\chain{\mathcal{M}}

\global\long\def\define{\stackrel{\Delta}{=}}

\global\long\def\argmin#1{\mathop{\textrm{argmin \,}}_{#1}}

\global\long\def\Norm#1{\Vert#1\Vert}
 \global\long\def\SqrNorm#1{\Vert#1\Vert^{2}}
 \global\long\def\Ltwo#1{\mathcal{L}^{2}\left(#1\right)}

\global\long\def\Normal#1#2#3{\mathcal{N}(#1;#2,#3)}

\global\long\def\LogNormal#1#2#3{ (#1-#2)^{T} #3^{-1} (#1-#2) }

\global\long\def\SqrMah#1#2#3{\Vert{#1}-{#2}\Vert_{#3}^{2}}

\global\long\def\SqrZMah#1#2{\Vert{#1}\Vert_{#2}^{2}}

\global\long\def\Info#1#2#3{\mathcal{N}^{-1}(#1;#2,#3)}

\providecommand{\half}{\frac{1}{2}} 

\global\long\def\Mah#1#2#3{\Vert{#1}-{#2}\Vert_{#3}}
 \global\long\def\MahDeriv#1#2#3#4{\biggl(\deriv{#2}{#4}\biggr)^{T} #3^{-1} (#1-#2)}

\global\long\def\argmin#1{\mathop{\textrm{argmin \,}}_{#1}}
 \global\long\def\argmax#1{\mathop{\textrm{argmax \,}}_{#1}}

\global\long\def\deriv#1#2{\frac{\partial#1}{\partial#2}}

\global\long\def\at#1#2{#1\biggr\rvert_{#2}}

\global\long\def\Jac#1#2#3{ \at{\deriv{#1}{#2}} {#3} }

\global\long\def\Rone{\mathbb{R}}
\global\long\def\Pone{\mathbb{P}}

\global\long\def\Rtwo{\mathbb{R}^{2}}
\global\long\def\Ptwo{\mathbb{P}^{2}}

\global\long\def\Stwo{\mathbb{S}^{2}}
 \global\long\def\Complex{\mathbb{C}}

\global\long\def\Z{\mathbb{Z}}
 \global\long\def\Rplus{\mathbb{R}^{+}}

\global\long\def\SOtwo{SO(2)}
\global\long\def\sotwo{\mathfrak{so(2)}}
\global\long\def\skew#1{[#1]_{+}}

\global\long\def\SEtwo{SE(2)}
\global\long\def\setwo{\mathfrak{se(2)}}
\global\long\def\Skew#1{[#1]_{\times}}

\global\long\def\Rthree{\mathbb{R}^{3}}
\global\long\def\Pthree{\mathbb{P}^{3}}

\global\long\def\SOthree{SO(3)}
\global\long\def\sothree{\mathfrak{so(3)}}

\global\long\def\Rsix{\mathbb{R}^{6}}
\global\long\def\SEthree{SE(3)}
\global\long\def\sethree{\mathfrak{se(3)}}

\global\long\def\Rn{\mathbb{R}^{n}}

\global\long\def\Afftwo{Aff(2)}
\global\long\def\afftwo{\mathfrak{aff(2)}}

\global\long\def\SLthree{SL(3)}
\global\long\def\slthree{\mathfrak{sl(3)}}

\global\long\def\stirling#1#2{\genfrac{\{}{\}}{0pt}{}{#1}{#2}}

\global\long\def\matlabscript#1#2{\begin{itemize}\item[]\lstinputlisting[caption=#2,label=#1]{#1.m}\end{itemize}}

\global\long\def\atan{\mathop{atan2}}

\global\long\def\degree{N}

\renewcommand{\bibfont}{\footnotesize} 


\begin{abstract}


We present a novel continuous time trajectory representation based on a Chebyshev polynomial basis, which when governed by known dynamics models, allows for full trajectory and robot dynamics estimation, particularly useful for high-performance robotics applications such as unmanned aerial vehicles.
We show that we can gracefully incorporate model dynamics to our trajectory representation, within a factor-graph based framework, and leverage ideas from pseudo-spectral optimal control to parameterize the state and the control trajectories as interpolating polynomials. This allows us to perform efficient optimization at specifically chosen points derived from the theory, while recovering full trajectory estimates.
Through simulated experiments we demonstrate the applicability of our representation for accurate flight dynamics estimation for multirotor aerial vehicles. The representation framework is general and can thus be applied to a multitude of high-performance applications beyond multirotor platforms.

\end{abstract}


\section{Introduction}

High-performance autonomous robotic platforms are increasingly important in a variety of complex and dangerous tasks which normally require expert human piloting or intervention. Tasks such as exploration and reconnaissance, inspection, precision agriculture, search and rescue, and autonomous camera platforms are examples where autonomy is being successfully leveraged, with more applications being unlocked as the state of the art in robotics improves.

State estimation is essential for robotic applications~\cite{Abeywardena13arxiv,Eckenhoff20tro}, but equally important is estimating the dynamics and control which induces the robot's trajectory.
In real-world scenarios, robots have to deal with highly dynamic conditions with noisy sensor information~\cite{Svacha19ral}.
Under such conditions, constraining the state estimation with known robot dynamics allows for more robust state estimation.
Conversely, the state estimates at any point on the trajectory can be used as a means to acquire knowledge about the dynamics at that time instance, such as estimating unmeasured force inputs.
Some examples of unmeasured forces include rotor and actuator forces/torques, forces due to adversarial conditions (e.g.\ strong winds), and contact forces with the ground or objects. An additional application is performing system identification to better estimate intrinsic and inertial parameters to allow for more precise robot control~\cite{Wuest19icra}.


While state estimation has seen considerable research interest, optimizing for the control estimates has not been tackled before, to the best of our knowledge.
Research on using robot kinematics to constrain state estimation has seen much interest in recent years. \citet{Hartley18icra_contact_factors} and~\citet{Wisth19ral_legged_fg} leverage the use of the forward kinematics from encoder measurements to estimate contact with the ground, for biped and quadraped robots respectively.
\citet{Nisar19rss_vimo} do model the dynamics of the robot in the optimization process via preintegration, however, they only consider linear forces measured from an IMU and do not estimate the actual control inputs.
Recent work has attempted to model the dynamical model parameters~\cite{Kaufmann20rss,Pan18rss} directly from sensor information using learning-based approaches. However, these tend to be brittle, poorly understood, and data and compute intensive, making general applicability difficult.


In this paper, we propose a novel mathematical framework for estimating both the state and control of a robot body, leveraging a pseudo-spectral parameterization based on Chebyshev polynomials. Our approach follows from the pseudo-spectral optimal control literature~\cite{Fahroo02gcd,Elnagar98coa} and utilizes a set of sparse collocation points along the trajectory within a factor graph based framework which allows for efficient optimization and inference, while providing significant accuracy and extensibility. The proposed framework only assumes knowledge of the robot's dynamics model, and we demonstrate the applicability of our approach on a quadrotor model to estimate both the states over a trajectory and also the rotor speeds and forces, using only measurement information from a monocular camera. In contrast to prior work, we do not make use of other sensors (such as IMUs) in the estimation pipeline. While we illustrate results on a quadrotor platform, our framework can be generally applied to a variety of robot models, as well as tackle various problems such as wind estimation, contact force estimation, system identification, etc. To the best of our knowledge, our proposed method is the first to leverage ideas from pseudo-spectral optimization for the problem of state estimation.





\section{Preliminaries}

\subsection{Chebyshev Polynomials and Chebyshev Points}
In this section we briefly review Chebyshev polynomials and their use in approximating continuous functions. We largely follow the exposition in \cite{Trefethen13book}, which thoroughly presents the numerical advantages of Chebyshev polynomials over other basis functions. We defer the numerical analysis in this work for the sake of brevity. The reader might also be interested in the hands-on development in \cite{Trefethen00book}.

\begin{figure}
\begin{centering}
\includegraphics[viewport=70bp 284bp 552bp 506bp,clip,width=1\columnwidth]{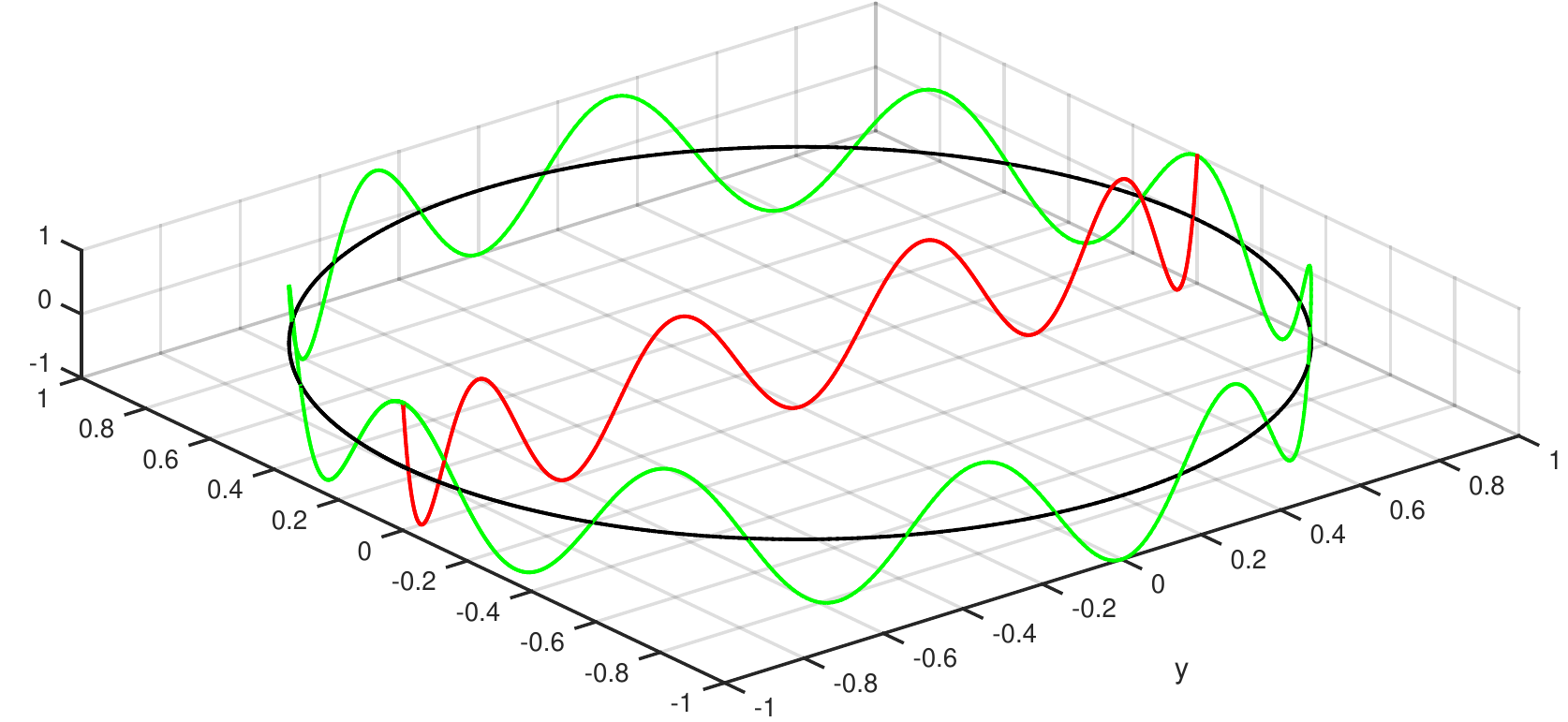}
\par\end{centering}

\begin{centering}
\includegraphics[viewport=70bp 290bp 552bp 510bp,clip,width=0.7\columnwidth]{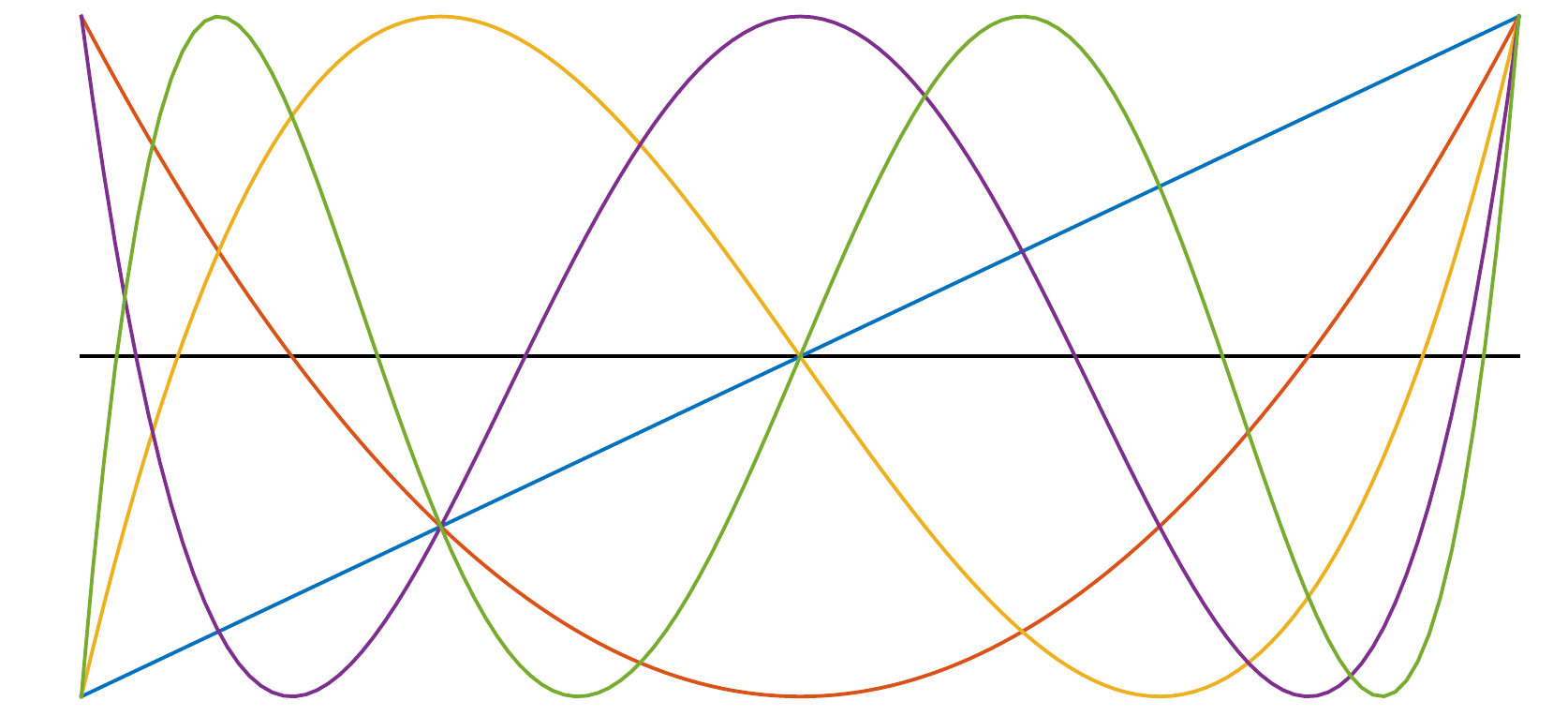}
\par\end{centering}

\centering{}\caption{\label{fig:T10}Top: Chebyshev polynomial $T_{10}$ of degree 10 (red), seen to be the projection of equidistant points from the unit circle (green) onto the x-interval $\left[-1,1\right]$.
Bottom: Chebyshev polynomials $T_{1}$ through $T_{5}$, alternatingly odd and even.}
\end{figure}

\global\long\def\tint{\tau}

The Chebyshev series is a spectral decomposition composed of an orthogonal basis defined on a unit circle, analogous to the Fourier series. It is defined for functions $f(\tint)$
on the interval $\left[-1,1\right]$, but can be easily scaled to arbitrary bounds. The series decomposition is defined as:
\begin{equation}
f(\tint)=\sum_{k=0}^{\infty}a_{k}T_{k}(\tint)\;\;\mbox{with}\;\;a_{k}=\frac{2}{\pi}\int_{-1}^{1}\frac{f(\tint)T_{k}(\tint)}{\sqrt{1-\tint^{2}}}\label{eq:series}
\end{equation}
with the factor $2/\pi$ changed to $1/\pi$ for $k=0$.

Above, $T_{k}$ is the $k$th Chebyshev polynomial\textbf{,} defined as the projection of a cosine function to the midline $\left[-1,1\right]$ of the unit circle, as shown in figure \ref{fig:T10}:
\begin{eqnarray}
\label{eq:Tk}
T_{k}(\tint) & \define & \cos\left(k\arccos(\tint)\right), \; -1 \leq \tint \leq 1
\end{eqnarray}

Given an arbitrary real function $f$ on $\left[-1,1\right]$, we can exploit the connection to the Fourier series by making use of the fast Fourier transform (FFT) to efficiently obtain the coefficients $a_{k}$ of the truncated Chebyshev series. To ensure convergence of the approximation and avoid issues such as Runge's phenomenon~\cite{Berrut04siam}, for a given degree of approximation $\degree$, we query $f$ at the following points:
\begin{equation}
\tint_{j}\define\cos\left(j\pi/\degree\right),~0\leq j\leq\degree\label{eq:chebpoints}
\end{equation}
\begin{figure}[h]
\centering{}\includegraphics[viewport=70bp 280bp 550bp 515bp,clip,width=0.7\columnwidth]{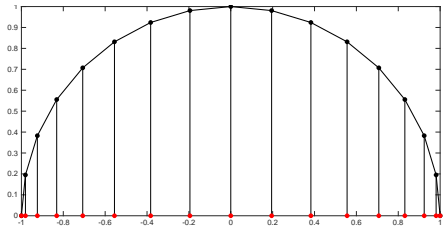}
\caption{\label{fig:Chebyshev-Gauss-Lobatto}Chebyshev points $\cos(k\pi/n)$ for degree $n=16.$ They are obtained by projecting a regular grid on the unit circle onto the x-axis. \emph{Figure generated using MATLAB code by L.N. Trefethen in \cite{Trefethen13book}.}}
\end{figure}

The points defined by (\ref{eq:chebpoints}) are the Chebyshev-Gauss-Lobatto (CGL) points \cite{Fahroo02gcd} or simply, Chebyshev points. Figure \ref{fig:Chebyshev-Gauss-Lobatto} shows that they arise simply from the projection of a regular grid on the unit circle.

The FFT of this function will only have $\degree$ non-zero values corresponding to cosines of increasing frequency, in one-to-one correspondence to the Chebyshev polynomials (\ref{eq:Tk}). Those non-zeroes are exactly the coefficients $a_{k}$, with $0\leq k\leq\degree$.

\subsection{Barycentric Interpolation}

Given the samples of $f$ at the $\degree+1$ Chebyshev points $f_j$, we can evaluate any point in $f$ efficiently using the Barycentric Interpolation formula~\cite{Berrut04siam}. The general Lagrangian form is given by:
\begin{equation}
f(x) = \sum_{j=0}^{\degree}\frac{\lambda_jf_j}{x-x_j}\Big/\sum_{j=0}^{\degree}\frac{\lambda_j}{x-x_j}
\end{equation}
where
\[
\lambda_j = \frac{1}{\Pi_{k \neq j}(x_j-x_k)}
\]
and $f(x) = f_j$ if $x = x_j$.

Computationally, the Chebyshev points provide an advantage over equidistant sampled points due to the simplicity of the resulting $\lambda$ values, giving us the following formula
\begin{equation}
\label{eq:barycentric}
   f(x) = \sum_{j=0}^{\degree}\frac{(-1)^jf_j}{x-x_j} \Big/ \sum_{j=0}^{\degree}\frac{(-1)^j}{x-x_j} 
\end{equation}
with the special case $f(x)=f_j$ if $x=x_j$, and the summation terms for $j=0$ and $j=\degree$ being multiplied by $1/2$.

Since this is a linear operation, we can reparameterize (\ref{eq:barycentric}) as an efficient inner product
$ f(x) = \textbf{f} \cdot \textbf{w} $
where $\textbf{f}$ is a vector of all the values of $f$ at the Chebyshev points, and $\textbf{w}$ is an $(\degree+1)$ vector of Barycentric weights.

\subsection{Differentiation Matrix}

Spectral collocation methods yield an efficient process for obtaining derivatives of the approximating polynomial via the differentiation matrix. Simply, an $\degree$ degree polynomial is determined by its values on the $(\degree+1)$ point grid, and its derivative is determined by its values on the same grid.

Thus, the derivatives of the interpolating polynomial can be efficiently computed via a matrix-vector product. This is useful when performing optimization as we can compute the derivatives of arbitrary functions for (almost) free.

\section{Approach}

\subsection{Problem Statement}

\global\long\def\measurement{\mathbf{z}}
\global\long\def\state{\mathbf{x}}
\global\long\def\control{\mathbf{u}}
\global\long\def\time{t}

We consider the following continuous-time optimization problem: determine the control function $\control(\time)$, and the corresponding state trajectory $\state(\time)$ at specific points, that jointly minimize a cost function of the form:
\begin{equation}
\sum_{i=1}^{m}\SqrMah{\measurement_{i}}{\mathbf{h}_{i}\left(\state(\time_{i}), \control(\time_{i}), \Theta_{i}\right)}{R_{i}}\label{eq:objective}
\end{equation}
where $\measurement_{i}$ is one of $m$ (vector-valued) measurements, $\state\left(\time_{i}\right)$ and $\control\left(\time_{i}\right)$ are the $n$-dimensional vehicle state and the $p$-dimensional control function respectively at the corresponding measurement time $\time_{i}$, $\mathbf{h}_{i}$ is a nonlinear measurement model, and $\Theta_{i}$ is a set of parameters that $\measurement_{i}$ also depends on, e.g., landmark coordinates, intrinsic parameters, etc.

Without loss of generality we assume Gaussian measurement noise, and denote the covariance matrix of the noise on $\measurement_{i}$ as $R_{i}$.
However, the formulation is easily extended to use other noise models, e.g., robust error norms. Priors on both the known variables $\Theta$ and the unknown variables can be easily accommodated as well.

We further assume that the vehicle is subject to the following continuous dynamics model,
\begin{equation}
\dot{\state}(\time)=\mathbf{f}\left(\state(\time),\control(\time)\right),~\time_{0}\leq\time\leq\time_{f},\label{eq:dynamics}
\end{equation}
with $\time_{0}\leq\time_{i}\leq\time_{f}$ for all discrete measurement times $\time_{i}$. For the sake of simplicity, we omit (in-)equality constraints as are customarily stated in optimal control including possible boundary conditions, which are easily taken into account.

In this paper we propose the use of Chebyshev polynomials to parameterize the continuous state trajectory $\state(\time)$ and the continuous control input $\control(\time)$, following the lead from the optimal control literature\cite{Elnagar98coa,Fahroo02gcd}, to render the variational optimization problem (\ref{eq:objective}) into a simple Non-Linear Programming (NLP) problem.

\subsection{Pseudo-spectral Parameterization}

The pseudo-spectral parameterization consists of $\degree$ function values $f_{j}$ at the Chebyshev points $\tint_{j}$. This allows us to estimate an interpolating polynomial whose values at the Chebyshev points are exactly the function values, while being smooth and efficiently differentiable.
Since we can easily switch between the function values $f_{j}$ and the series coefficients $a_{k}$ using the FFT, or indeed back again using the inverse FFT, the two representations are equivalent.

\begin{figure}[h]
\centering{}\includegraphics[viewport=50bp 260bp 570bp 510bp,clip,width=1\columnwidth]{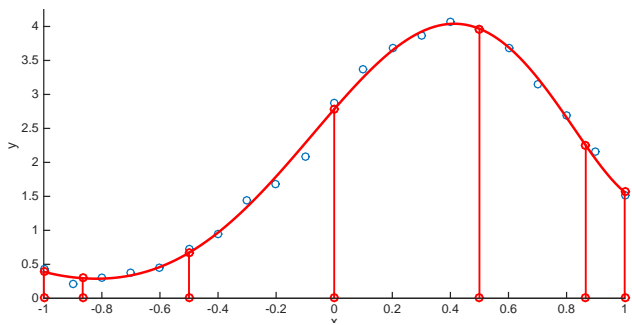}
\caption{\label{fig:Chebyshev2} Least-squares fit of a degree 6 Chebyshev interpolant to 21 noisy samples of the function $\exp(\sin(2x)+\cos(2x))$.
The parameters we optimize for are the 7 values at the Chebyshev-Gauss-Lobatto points, indicated by the stem plots. Note they do not in general coincide with any of the samples.}
\end{figure}

The idea is illustrated using a scalar example in Figure \ref{fig:Chebyshev2}, where we fit a degree 6 polynomial to 21 noisy samples of a known function.
The parameters are the 7 values of the sought function at the $6+1$ Chebyshev points. For every combination of these 7 values, there exists a unique polynomial interpolant $I(\tint)$, for $-1\leq\tint\leq1$.
The optimal values minimize the squared distance between $I(\tint)$ and the samples. We discuss below exactly how this minimization can be achieved in the general optimization setting.

\subsection{Approximating Trajectories and Control Functions}

\global\long\def\states{\mathbf{X}}
\global\long\def\controls{\mathbf{U}}

We borrow the main idea from pseudo-spectral optimal control by parameterizing the unknown, continuous state and control functions using their values at the Chebyshev points, i.e., using a pseudo-spectral parameterization. 

Arbitrary time intervals $\left[\time_{0},\time_{f}\right]$ can be accommodated using the affine transformation \cite{Fahroo02gcd}
\[
\time=\left[(\time_{f}+\time_{0})+(\time_{f}-\time_{0})\tau\right]/2.
\]
With this transformation, we define the \emph{pseudo-spectral state parameterization} as the $\degree$-dimensional vectors $\mathbf{x}_{s}$, one for each of the $m$ state variables $x_{s}(\time)$. These represent the state values at the transformed Chebyshev points
\begin{equation}
\time_{j}\define\frac{\time_{f}+\time_{0}}{2}+\frac{\time_{f}-\time_{0}}{2}\cos\left(j\pi/\degree\right),~0\leq j\leq\degree.\label{eq:transformed}
\end{equation}

Similarly, we define the \emph{pseudo-spectral control parameterization} as the $\degree$-dimensional vectors for each of the $p$ control variables. We collect all the parameters in the $m\times(\degree+1)$ matrix $\states$ and the $p\times(\degree+1)$ matrix $\controls$, which constitute our unknowns in what follows.


\subsection{Pseudo-spectral Optimization: Minimizing Measurement Error }

The key idea is to express the desired objective function (\ref{eq:objective}) in terms of the pseudo-spectral parameterization $\{\states, \controls \}$. To this end, we replace the optimal control performance index with the sum of measurement-derived least-squares terms, while keeping the idea of enforcing the dynamic constraints through collocation.

\global\long\def\weights{\mathbf{w}}
\global\long\def\wi{\mathbf{\weights_{i}}}
\global\long\def\deltaX{\mathbf{\Delta}}

The main mechanism we will use is the barycentric interpolation formula~\ref{eq:barycentric} to predict the state $\state(\time)$ from $\states$ at any arbitrary time $\time$. For a given $\time$, trajectory interpolation can be written as
\begin{equation}
\label{eq:state}
    \state(\time) = \states \weights(\time)
\end{equation}
with $\weights(\time)$ being an $(\degree+1)$-dimensional weight vector. Substituting (\ref{eq:state}) in the objective function (\ref{eq:objective}), and writing $\wi=\weights(\time_{i})$ we obtain 
\begin{equation}
\label{eq:states}
    E_{1}(\states)=\sum_{i=1}^{m}\SqrMah{\measurement_{i}}{\mathbf{h}_{i}\left( \states\wi \right)}{R_{i}}
\end{equation}
which we can minimize using non-linear programming.

\global\long\def\DN{\mathbf{D}_{\degree}}

Of particular note is the ease by which measurements on \emph{derivatives} of the state can be accommodated. We have 
\begin{equation}
    \label{eq:diff}
    \dot{\state}(\time) = \DN \states \weights(\time)
\end{equation}
where $\DN$ is an $(\degree+1)\times(\degree+1)$ differentiation matrix, defined in \cite[p. 53]{Trefethen00book}.
The equation above can be substituted in the measurement equation in an analogous manner. Similar matrices can be defined for the second time derivative, etc.

\subsection{Enforcing Dynamics through Direct Collocation}

The final piece in the puzzle is enforcing the vehicle dynamics, which
we do through direct collocation, similar to optimal
control~\cite{Hargraves87gcd,vonStryck93chapter}. This method optimizes over both state and controls while enforcing the dynamics constraints at a set of ``collocation points'' as dynamic defect terms, and is not limited to pseudospectral methods.

\global\long\def\wj{\mathbf{\weights_{j}}}

To this end, we similarly write the continuous control function $\control(\time)$
as a linear function of the parameters $\control(\time) = \controls \weights(\time)$,
and substituting (\ref{eq:state}) and (\ref{eq:diff}) into
the dynamics equation (\ref{eq:dynamics}), we obtain 
\begin{equation}
    \label{eq:defects}
    \DN\states\wj = \mathbf{f}\left(\states\wj,\controls\wj,\time_{j}\right),~0\leq j\leq\degree
\end{equation}
which is a hard, possibly nonlinear, equality constraint at each of the
(transformed) Chebyshev points $\time_{j}$, with $\wj=\weights(\time_{j})$.
A stochastic version of the same would amount to minimizing the following
least-squares objective:
\begin{equation}
\label{eq:stochastic}
    E_{2}\left(\states,\controls\right) \define \sum_{j=0}^{\degree} \SqrMah{\DN\states\wj}{\mathbf{f}\left(\states\wj,\controls\wj,\time_{j}\right)}{Q}
\end{equation}
with $Q$ an $m \times m$ covariance matrix, which besides stochasticity
can also account for model error, and where we assume a Gaussian noise
model for simplicity.

Our approach assumes the availability of the robot dynamics in the control-affine form, either in an analytical form or as a learned model. Our optimization-based approach allows for minor inaccuracies in the dynamics model (such as environmental factors) and is robust to noisy initial estimates. We leave a thorough analysis of the dynamics model and its effects on the final estimates as potential future work.

\subsection{Nonlinear Programming}

The final objective function to be minimized is thus
\begin{equation}
\label{eq:sum}
    E\left(\states,\controls\right)\define E_{1}\left(\states\right)+E_{2}\left(\states,\controls\right)
\end{equation}
which is the sum of measurement error, interpolated at $T$ arbitrary
time instants $\time_{i}$, $1\leq i\leq T$, and the dynamic defects
at the $\degree+1$ transformed Chebyshev points $\time_{j}$, $0\leq j\leq\degree$.

One last detail to be dispensed with is how to minimize over the \emph{matrices}
of state values $\states$ and control values $\controls$, as most
NLP software packages expect to iteratively update perturbation \emph{vectors}.
In detail, we linearize the measurement function $\mathbf{h}_{i}$,
which is done through the Taylor expansion,
\[
\mathbf{h}_{i}\left[\left(\bar{\states}+\deltaX\right)\wi\right] \approx \mathbf{h}_{i}\left(\bar{\states}\wi\right)+\mathbf{H_{i}}\deltaX\wi
\]
where $\deltaX$ is an $m\times(\degree+1)$ matrix and $\mathbf{H}_{i}$ is the measurement Jacobian of $\mathbf{h}_{i}$
with respect to the state $\state$. To obtain a vector-valued perturbation on the estimate
$\bar{\states}$ we can rewrite
\[
\mathbf{H}\deltaX\wi=\left( \wi^{T} \varotimes \mathbf{H}_{i} \right) \delta_{\state}
\]
where $\varotimes$ is the standard Kronecker product, and $\delta_{\state}=\mathop{vec}(\deltaX)$
is the $m (\degree+1)$-dimensional perturbation vector obtained by
stacking the columns of $\deltaX$. Note that in general the matrix
$\wi^{T}\varotimes\mathbf{H}_{i}$ is a dense matrix, unless $\mathbf{h_{i}}$
does not depend on one or more state variables.

Similarly, we can linearize the nonlinear dynamics \textbf{$\mathbf{f}$} as
\begin{eqnarray*}
 &  & \mathbf{f}\left[\left(\bar{\states}+\deltaX\right)\wj,\left(\bar{\controls}+\Sigma\right),\time_{j}\right]\\
 & \approx & \mathbf{f}\left(\bar{\states}\wj,\bar{\controls}\wj,\time_{j}\right)+\mathbf{F}_{j}\deltaX\wj+\mathbf{G}_{j}\Sigma\wj\\
 & = & \mathbf{f}\left(\bar{\states}\wj,\bar{\controls}\wj,\time_{j}\right)+\left(\wj^{T}\varotimes\mathbf{F}_{j}\right)\delta_{x}+\left(\wj^{T}\varotimes\mathbf{G}_{j}\right)\delta_{u}
\end{eqnarray*}
where we defined $\delta_{\control}=\mathop{vec}(\Sigma)$ as the
$p(\degree+1)$-dimensional perturbation vector on the controls estimate
$\bar{\controls}$. The linearization coupled with the sparsity of the Chebyshev points and the linear Barycentric Interpolation formula allows the use of standard linear solvers to achieve an efficient solution.

\section{Application to Quadrotors}

Quadrotors are an important class of autonomous agents used in a variety of applications, such as surveillance and transport, due to their speed and versatility. This has lead to increased recent research on various areas involving quadrotors, such as state estimation, parameter estimation, and control \cite{Abeywardena13arxiv,Svacha19ral,Nisar19rss_vimo,Eckenhoff20tro,Wuest19icra}.


\subsection{Quadrotor Dynamics}
We follow~\textcite{Beard08report} and~\textcite{Altug05ijrr} to describe the quadrotor dynamics. We assume an X-configuration quadrotor as is standard in the literature, with the motors numbered counter-clockwise starting from the top-left when looking at the quadrotor from above. We denote the mass of the quadrotor as $m$, inertial tensor as $\mathbf{I}$, and vector of motor speeds as $u$. We denote the world frame with $n$ (for navigation frame as is common in the aerospace literature) and the body-frame of the vehicle as $b$.

The Newton-Euler dynamics equations are given by
\begin{equation}
m\dot{\mathbf{v}}^n = \mathbf{F}^n
\end{equation}
\begin{equation}
\mathbf{I}\dot{\boldsymbol{\omega}}^b = \boldsymbol{\tau}^b - \boldsymbol{\omega}^b \times \mathbf{I}\boldsymbol{\omega}^b
\end{equation}

\subsubsection{Force}

The total force acting on the center of mass of the quadrotor is~\cite{Altug05ijrr,Castillo05book}
\begin{equation}
\mathbf{F}^n = m\mathbf{g}^n + \mathbf{k}^n_b \sum{k_fw_i^2} + \mathbf{f}_{d}
\end{equation}
where the first term is due to the Earth's gravitational force, $\mathbf{k}^n_b$ is the body z-axis in the world frame, $w_i$ is the speed of the $i$th motor, $k_f$ is the thrust coefficient, and $\mathbf{f}_d = k_d \Vert v^n \Vert v^n$ is the force due to aerodynamic drag with $k_d$ being the drag constant assuming the density of air is constant.

\subsubsection{Torque}

We can represent the torque cross product as a matrix multiplication via a skew-symmetric matrix
\begin{equation}
\tau = 
\mathbf{I}
\dot{\boldsymbol{\omega}^b}
+ \\
\begin{bmatrix}
0 & -r & q \\
r & 0 & -p \\
-q & p & 0 \\
\end{bmatrix}
\mathbf{I}
\boldsymbol{\omega}^b
\end{equation}
where
$$
\boldsymbol\omega^b = 
\begin{bmatrix}
p \\
q \\
r
\end{bmatrix},
\mathbf{I} = \begin{bmatrix}
I_{xx} & 0 & 0 \\
0 & I_{yy} & 0 \\
0 & 0 & I_{zz}
\end{bmatrix}
$$
are the angular velocity and the inertial tensor respectively.
Thus, we get the angular moments as,
\begin{equation}
\begin{bmatrix}
	I_{xx}\dot{p} \\
	I_{yy}\dot{q} \\
	I_{zz}\dot{r} \\
\end{bmatrix}
= 
\begin{bmatrix}
	\tau_x \\
	\tau_y \\
	\tau_z \\
\end{bmatrix} 
- 
\begin{bmatrix}
	qr I_{zz} - qr I_{yy} \\
	pr I_{xx} - pr I_{zz} \\
	pq I_{yy} - pq I_{xx}\\
\end{bmatrix}
\end{equation}
The total external torque (moments) applied in the body frame is given by,
\[
\boldsymbol\tau^B = \boldsymbol\tau^T - \boldsymbol{g}_a + \boldsymbol\tau_{w}
\]
where $\boldsymbol\tau^T$ is the torque due to the motor speeds, $g_a$ is the gyroscopic moments, and $\boldsymbol\tau_{w}$ is the torque due to drag. The gyroscopic moments are considered negligible~\cite{Mahony12ram}.

Given the motor configuration for the quadrotor, and the motor arm length $l$, we get $\boldsymbol\tau^T$ via a mixing matrix

\begin{equation}
\begin{bmatrix}
	\tau_x \\
	\tau_y \\
	\tau_z
\end{bmatrix}
=
\begin{bmatrix}
	l & l & -l & -l \\
	-l & l & l & -l \\
	-C & C & -C & C
\end{bmatrix}
\begin{bmatrix}
	f_1 \\
	f_2 \\
	f_3 \\
	f_4
\end{bmatrix}
\end{equation}

Here, $f_i$ is the thrust force of each motor, and $C$ is the ratio of torque coefficient to thrust coefficient. The signs in the first 2 rows of the mixing matrix depend on the vehicle frame, and the signs in the third row depend on the direction of the motor rotation, with positive yaw direction being negative.

\subsection{Time Derivatives}

Given the dynamics equation~(\ref{eq:dynamics}), we can compute the Jacobian matrix $J$ by computing the time derivative for each term of the 12 dimensional state vector.

\subsubsection{Position}
This is simply the velocity, $\dot{\mathbf{p}} = \mathbf{v}$.

\subsubsection{Rotation}
Our parameterization of rotation is based on the Lie Algebra $so(3)$ to allow for optimization. By Euler's Theorem, this implies that the time derivative of the rotation is the angular velocity in the body frame, $\dot{\mathbf{R}} = \boldsymbol{\omega}$.

\subsubsection{Velocity}
This is the normalized total force, $\dot{\mathbf{v}} = \mathbf{f}/m$.

\subsubsection{Angular Rate}
For the angular rate, the time derivative is simply the inertial tensor normalized torque, $\dot{\boldsymbol{\omega}} = \mathbf{I}^{-1}\boldsymbol{\tau}$

%

\subsection{Experimental Setup}

For collecting measurements and evaluating our proposed framework, we leverage the \textit{FlightGoggles} simulator~\cite{Guerra19iros}. FlightGoggles provides us with both a photo-realistic and dynamically accurate simulation environment, as well as easy querying of quadrotor state \& dynamics in real-time for data collection with some minor modifications. We make the standard configuration for the simulated quadrotor and all the default simulator parameters.

Our proposed framework is based on the GTSAM factor graph optimization library~\cite{Dellaert17fnt_fg}. The measurement error is computed using the projection factors available in GTSAM. To model the dynamics defects, we define a new type of factor which takes as input the state and control matrices $\{\mathbf{X}, \mathbf{U}\}$ respectively. All our experiments use $\degree=128$ for the polynomial degree, which balances efficiency and generality, eliminating the need for manual tuning.

For measuring initial estimates of poses from images, we use a monocular visual odometry pipeline built upon GTSAM. We follow the standard procedure of performing feature matching and estimating the pose from the fundamental matrix. Note that we only use the image data for our experiments, however use of the IMUs can also be made to provide initial estimates for linear and angular velocities.

\begin{figure}[ht]
	\centering
	\includegraphics[width=\columnwidth]{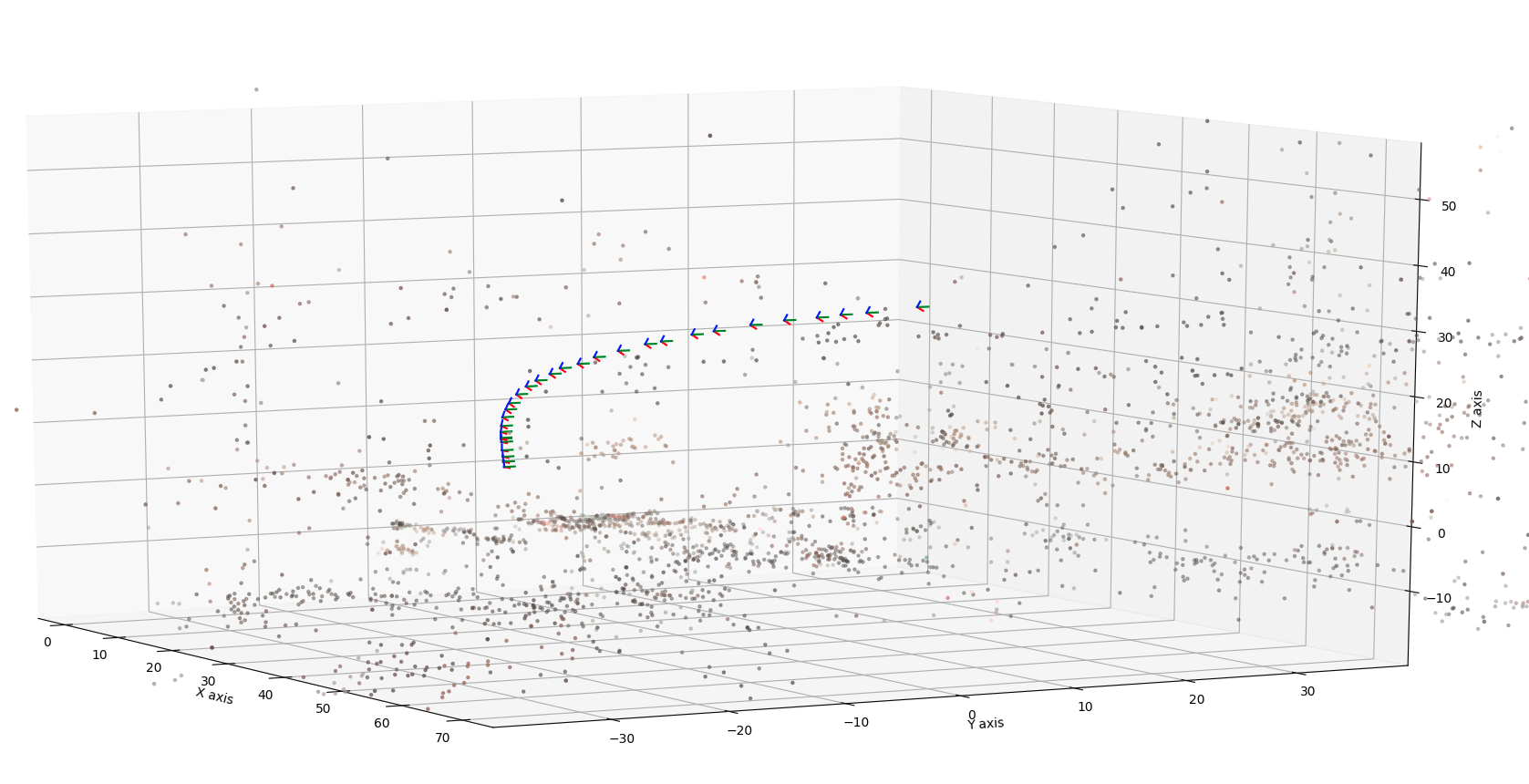}
	\caption{Example of an estimated state trajectory for a quadrotor platform. The axes represent the quadrotor poses and the points represent landmarks estimated from the visual odometry. We can estimate the trajectory as well as the control inputs using information from only a single camera sensor.}
	\label{fig:state-matrix-optimization}
\end{figure}

\section{Results}

\begin{figure*}[h]
	\centering
	\includegraphics[width=2\columnwidth]{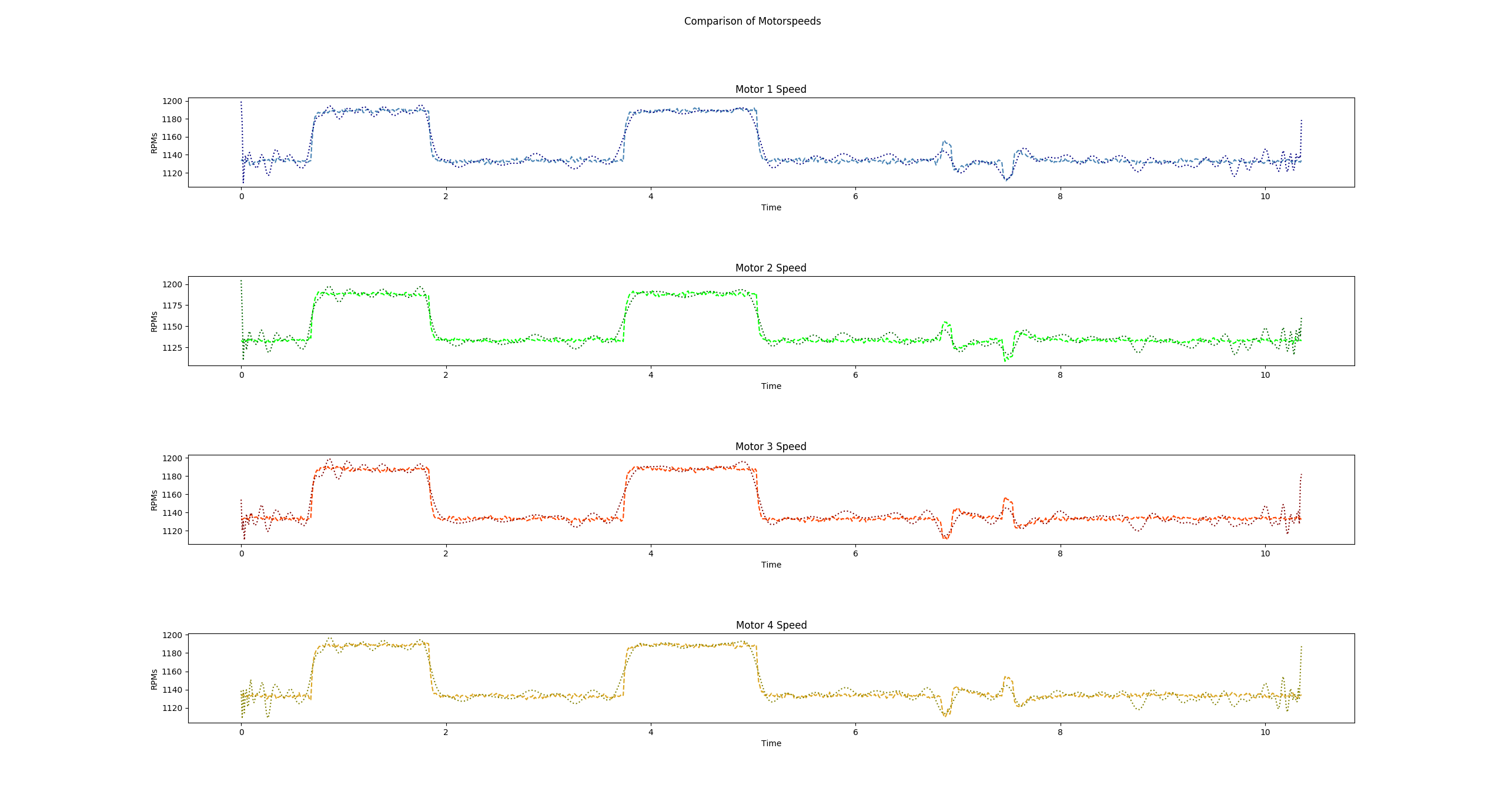}
	\includegraphics[width=2\columnwidth]{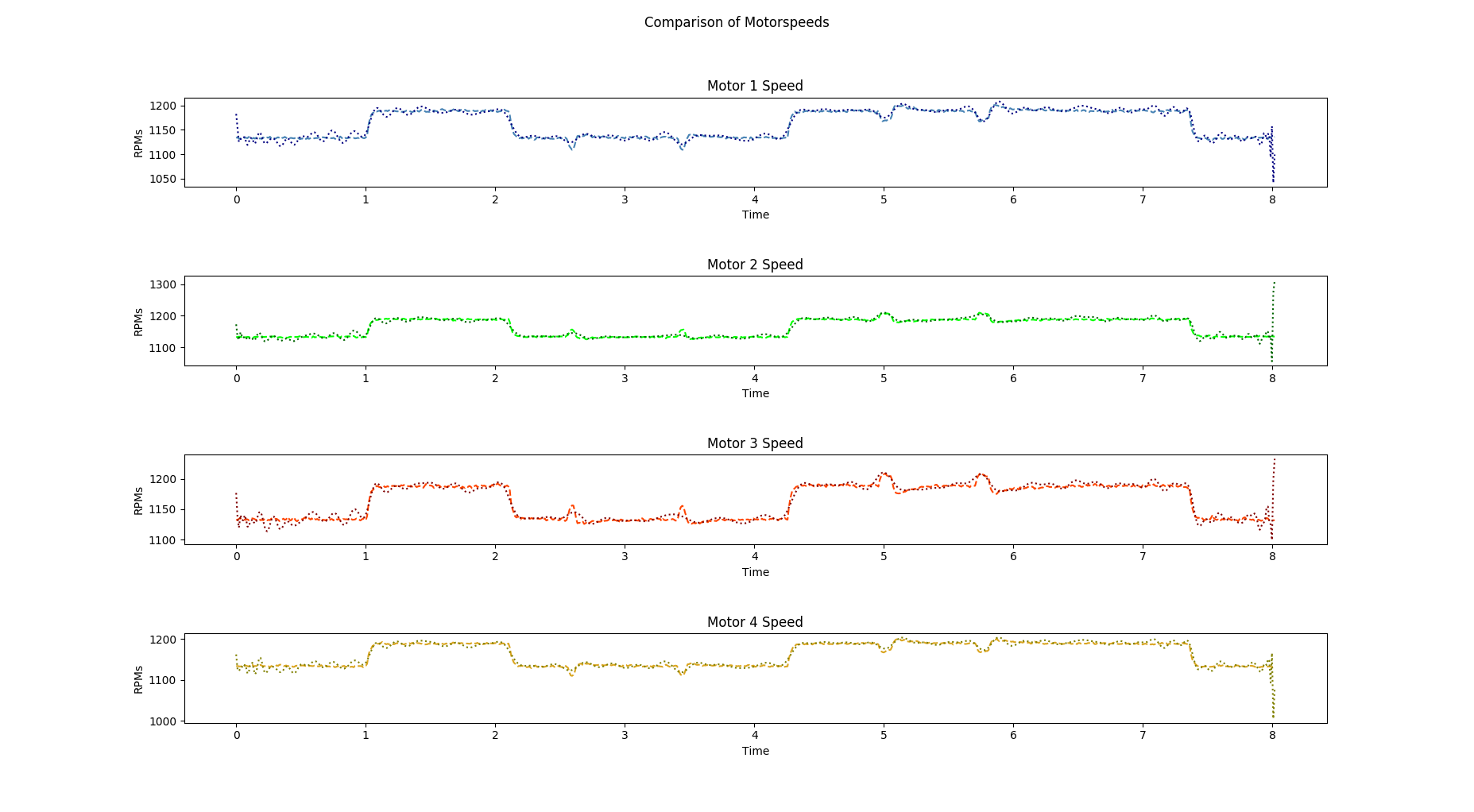}
	\caption{Comparison of motorspeeds for the actuators for 2 simulation runs. The dotted lines represent the ground truth and the dashed lines the estimates. Due to the global nature of the Chebyshev polynomials, we only see general trends rather than detailed resolution.}
	\label{fig:comparison_of_motorspeeds}
	\vspace{-1.5em}
\end{figure*}

We run $R=6$ different runs of the simulator, with increasing trajectory complexity, and estimate the state and control trajectories for each run. In terms of runtime efficiency, our state estimator runs in $\sim3.9$ minutes for a 565 state trajectory on a standard Linux desktop. This can be further optimized via the use of specialized matrix libraries, parallel processing, and polynomial degree / optimizer tuning.

A novel aspect of our estimator is that we can recover the dynamics inputs alongside the state. This is demonstrated in figure~\ref{fig:comparison_of_motorspeeds} which shows a qualitative comparison of the motor speeds. While the exact resolutions are difficult to see, the general trends are clearly observable, demonstrating our framework's capability.

The quantitative analysis of the estimated motorspeeds with respect to the ground truth provides more insight. We take the average error across the entire trajectory for each motor for each run of the simulator. The results are presented in table~\ref{tab:motorspeed-comparisons}, with lower values being better.

\begin{table}[!htb]
	\label{tab:motorspeed-comparisons}
	\centering
	\begin{tabular}[t]{| p{0.63cm} | p{0.63cm} | p{0.63cm} | p{0.63cm} | p{0.63cm} | p{0.63cm} | p{0.63cm} | p{0.63cm}|}
		\hline
		\multicolumn{2}{|c|}{\textbf{Motor 1}}  & \multicolumn{2}{c|}{\textbf{Motor 2}} & \multicolumn{2}{c|}{\textbf{Motor 3}} & \multicolumn{2}{c|}{\textbf{Motor 4}} \\ 
		\hline
		\hline
		RPM & \% err & RPM & \% err & RPM & \% err & RPM & \% err \\ 
		\hline
		1.7069 & 0.1441 & 1.9419 & 0.1641 & 1.9735 & 0.1667 & 1.6878 & 0.1425 \\
		\hline
		4.1462 & 0.3616 & 4.3242 & 0.3772 & 4.3716 & 0.3814 & 4.1131 & 0.3588 \\
		\hline		
		4.6440 & 0.3995 & 4.5240 & 0.3892 & 4.5707 & 0.3932 & 4.5167 & 0.3885 \\
		\hline 
		6.3759 & 0.5561 & 5.8155 & 0.5072 & 6.2201 & 0.5424 & 5.8438 & 0.5096 \\
		\hline
		7.0094 & 0.5919 & 7.3957 & 0.6240 & 7.0751 & 0.5977 & 7.3921 & 0.6240 \\
		\hline
		6.9770 & 0.5989 & 7.0844 & 0.6081 & 6.8208 & 0.5861 & 6.9810 & 0.5998 \\
		\hline
	\end{tabular}
	\caption{Average error between true and estimated motorspeeds.}
	\vspace{-2.5em}
\end{table}
As can be seen from the quantitative results in table~\ref{tab:motorspeed-comparisons}, our proposed framework is able to accurately estimate the control inputs with less than 1\% relative error.












\section{Conclusion}

In this paper, we have proposed a novel parameterization of the state and control trajectories based on Chebyshev polynomials in a pseudo-spectral optimization framework.
Our approach general, capable of being applied to various systems and robots without any major assumptions other than the dynamics model be known. Moreover, we achieve the estimates using only a single monocular camera, allowing for applicability in a wide range of applications, with potential for improvements by incorporating measurements from additional sensors such as IMUs. While we have demonstrated results in simulation for the purposes of quantitative analyses, the immediate next step would be to enable real-world deployment on robotic hardware, including, but not limited to, mobile manipulators and soft robots.

Another avenue of research would be to formulate our approach in incremental estimation settings, rather than the current global parameterization. This would allow our approach to be run in real-time and allow for online estimation of states and parameters, amongst other things. Finally, our framework could also be used in learning-based systems (e.g.\ imitation learning~\cite{Osa18ftr_imitation}) to allow for learning of complex dynamics which are hard to model manually. Our hope is that this work leads to further generalization of robots in real-world settings, and allows for easy and efficient estimation of robot parameters.





\newpage
\printbibliography

\end{document}